\documentclass[conference]{IEEEtran}
\usepackage{cite}
\ifCLASSINFOpdf
   \usepackage[pdftex]{graphicx}
\else
\fi
\usepackage{fixltx2e}
\usepackage{stfloats}

\begin{document}
%
% paper title
% Titles are generally capitalized except for words such as a, an, and, as,
% at, but, by, for, in, nor, of, on, or, the, to and up, which are usually
% not capitalized unless they are the first or last word of the title.
% Linebreaks \\ can be used within to get better formatting as desired.
% Do not put math or special symbols in the title.
\title{Churn Prediction in Mobile Social Games:\\Towards a Complete Assessment Using \\Survival Ensembles}

% author names and affiliations
% use a multiple column layout for up to three different
% affiliations

% \author{
% 	\IEEEauthorblockN{Authors' details omitted for double-blind reviewing}
% }

% \author{\IEEEauthorblockN{\'{A}frica Peri\'a\~{n}ez}
% \IEEEauthorblockA{Silicon Studio\\
% 1-21-3 Ebisu Shibuya-ku\\
% Tokyo, Japan\\
% Email:  africa.perianez@siliconstudio.co.jp}
% \\
% \IEEEauthorblockN{Anna Guitart}
% \IEEEauthorblockA{Silicon Studio\\
% 1-21-3 Ebisu Shibuya-ku\\
% Tokyo, Japan\\
% Email: anna.guitart@siliconstudio.co.jp}
% \and
% \IEEEauthorblockN{Alain Saas}
% \IEEEauthorblockA{Silicon Studio\\
% 1-21-3 Ebisu Shibuya-ku\\
% Tokyo, Japan\\
% Email: alain.saas@siliconstudio.co.jp}
% \\
% \IEEEauthorblockN{Colin Magne}
% \IEEEauthorblockA{Silicon Studio\\
% 1-21-3 Ebisu Shibuya-ku\\
% Tokyo, Japan\\
% Email: colin@siliconstudio.co.jp}
% }

\author{\IEEEauthorblockN{\'{A}frica Peri\'a\~{n}ez, Alain Saas, Anna Guitart and Colin Magne}
\IEEEauthorblockA{
Game Data Science Department\\
Silicon Studio\\
1-21-3 Ebisu Shibuya-ku, Tokyo, Japan\\
\{africa.perianez, alain.saas, anna.guitart, colin\}@siliconstudio.co.jp}
}
\maketitle

% As a general rule, do not put math, special symbols or citations
% in the abstract
\begin{abstract}
Reducing user attrition, i.e. churn, is a broad challenge faced by several industries. In mobile social games,
decreasing churn is decisive to increase player retention and rise revenues. Churn prediction models
allow to understand player loyalty and to anticipate when they will stop playing a game. Thanks to these
predictions, several initiatives can be taken to retain those players who are more likely to churn.

Survival analysis focuses on predicting the time of occurrence of a certain event, churn in our case.
Classical methods, like regressions, could be applied only when all players have left the game. The challenge 
arises for datasets with incomplete churning information for all players, as most of
them still connect to the game. This is called a censored data problem and is in the nature of churn.
Censoring is commonly dealt with survival analysis techniques, but due to the inflexibility of the
survival statistical algorithms, the accuracy achieved is often poor. In contrast, novel ensemble learning 
techniques, increasingly popular in a variety of scientific fields, provide high-class prediction results.

In this work, we develop, for the first time in the social games domain, a survival ensemble model which
provides a comprehensive analysis together with an accurate prediction of churn. For each player, we
predict the probability of churning as function of time, which permits to distinguish various levels of loyalty
profiles. Additionally, we assess the risk factors that explain the predicted player survival times. Our results show
that churn prediction by survival ensembles significantly improves the accuracy and robustness of
traditional analyses, like Cox regression.
\end{abstract}

% no keywords

% For peer review papers, you can put extra information on the cover
% page as needed:
% \ifCLASSOPTIONpeerreview
% \begin{center} \bfseries EDICS Category: 3-BBND \end{center}
% \fi
%
% For peerreview papers, this IEEEtran command inserts a page break and
% creates the second title. It will be ignored for other modes.
\IEEEpeerreviewmaketitle

\begin{figure}[ht!]
  \centering
  \includegraphics[width=0.49\columnwidth]{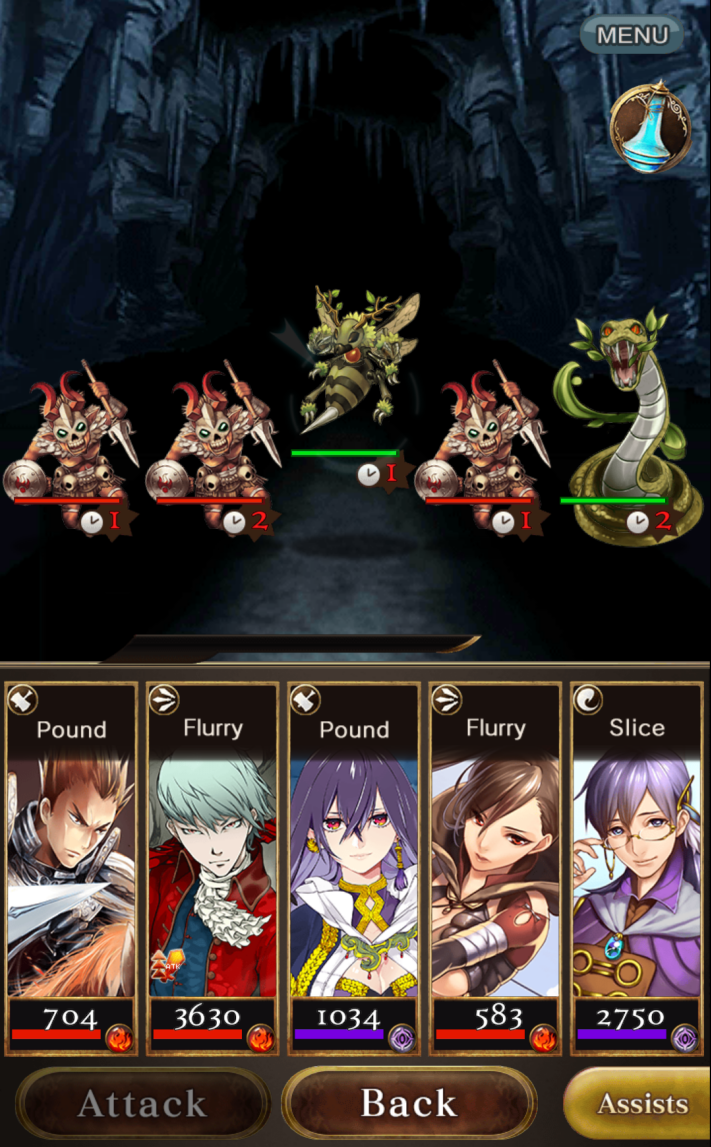}
  \includegraphics[width=0.482\columnwidth]{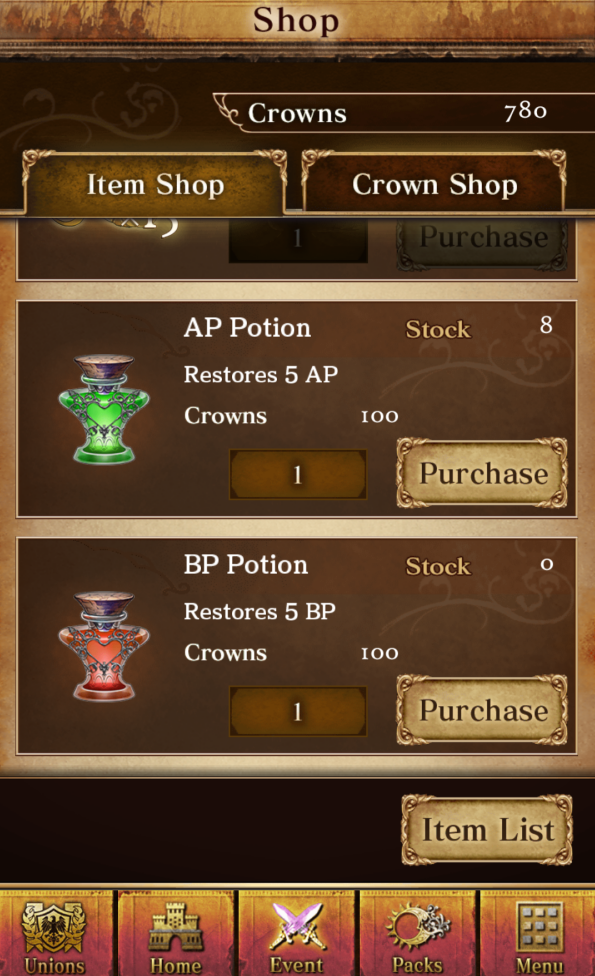} 
\caption{Screenshots of the game chosen to evaluate the churn models, Age of Ishtaria. This game is representative of the successful F2P mobile social role-playing games in Japan. The left panel exhibits characteristic activity in role-playing games, the right panel 
shows the usual F2P in-app purchases for in-game content.}
\label{ishtariaGame}
\end{figure}

\section{Introduction}
\label{intro}
% no \IEEEPARstart
The economics of gaming has changed in the recent years with the widespread adoption of social networks 
and smartphones, leading to a new type of video games: social games. Social games target a 
new audience of players: casual gamers, with a new monetization model: free-to-play (F2P or freemium), 
which now largely dominates all the mobile platforms \cite{Monetization,annie}. The freemium model consists in offering a game for free, and monetizing it by charging for in-game content through in-app purchases.

For social games, player retention is key for a successful monetization, and to increase the social interactions
that in turn help to drive the adoption of the game and retain players. In addition, the cost of acquiring 
new players is ever increasing \cite{Monetization} and can significantly exceed the cost of retaining existing ones.

This study is motivated by the idea that the ability to predict when a player will leave a game allows to take incentive actions to re-engage her and prevent churn, or move her to another game of the company.

Churn prediction has been widely researched in the fields of telecom, finance, retail, pay TV and banking, as shown by the extensive literature review given by \cite{verbeke2011building, tamaddoni2010modeling}. It has also been studied in e-commerce \cite{yoon2010prediction, yu2011extended} and even in terms of employee retention \cite{saradhi2011employee}.

In the field of video games, pioneering studies were introduced in \cite{ding,kawale}.
However, they focus on MMORPG (Massively Multiplayer Online Role-Playing Games).

MMORPG have been the first successful type of online social games, however they targeted a narrower audience and they are mainly using a subscription-based monetization model. This implies the possibility to measure churn as a formal termination of contract, similarly to the sectors mentioned above, at the exception of e-commerce.

Free-To-Play (F2P) monetization, which is the main model used by mobile social games, involves a non-contractual relationship. In this context, churn is not clearly determined by an explicit statement ending a contract. For the most active players, we can define churn as a prolonged period of inactivity. However, the problem slightly differs from the churn in e-commerce. It is indeed always possible for inactive users to come back to an e-commerce website, while inactive mobile players can uninstall a game, which would correspond to a well defined and definitive state of churn. However, this information is normally not available.
The definition of churn in non-contractual settings has been discussed in \cite{clemente2014methodology}. A comprehensive discussion on the definition of churn for F2P applications is beyond the scope of this paper, and is the subject of dedicated studies \cite{clemente2014methodology}.

The work presented in \cite{hadiji} is the first study investigating churn prediction in F2P games. \cite{hadiji} introduces a general definition of the problem, a selection of game content independent features and a comparison of classifiers. A second study shown in \cite{runge2014} focuses on the churn prediction of high value players in F2P games. \cite{runge2014} investigates in detail the problem definition and classifier evaluation, though it approaches the problem only from a binary classification point of view. It uses an algorithm that assumes a distribution of data that normally does not fit with the common shape of the churn data. Going further, \cite{saasTS2016_COMMENTED} and \cite{rothenbuehler2015hidden} try to address the temporality of the data for churn prediction in mobile games.
\\

The work presented in this paper focuses on predicting churn for {\em high value players} who are commonly called {\em whales} in the video game industry. A motivation for this focus is that whales behave differently than average players, including in terms of survival curve as we can see in Fig. \ref{KM}.
Since they are often the most active players, i.e. they play nearly every day, we can easily define their churn as a prolonged period of inactivity. Their high level of engagement also allows to collect more data about their activity and makes them more likely to answer positively to actions taken in order to prevent their churn. Finally, from a business perspective, whales, who represent about 0.15$\%$ of the players, or 10$\%$ of the paying users \cite{whales}, are particularly important since they are the top spenders who account for 50$\%$ of the in-app purchases revenues.
\\

The game chosen for this study, Age of Ishtaria developed by Silicon Studio, is representative of the successful mobile social games and has several million players worldwide.

%\hfill mds
 
%\hfill August 26, 2015

\subsection{Our contribution}
Classical approaches to churn study the problem as a binary classification: whether or not the player connects again to the game (e.g. \cite{banasik1999not}). Although the 
binary models are very intuitive, they are not able to predict {\em when} 
the player will stop playing and, moreover, the features are limited to provide static (non-temporal) information.

In order to model the {\em time} until {\em churn}, traditional methods like regressions would be appropriate only when all players have stopped playing the game. The challenge arises for 
data which contains incomplete information about every users, as some of them still play the game. 

The present work improves previous studies \cite{hadiji,runge2014} using an adequate technique that assimilates censored data (observations with incomplete information about churn time) \cite{lagakos1979general} and that captures the temporal dimension of the churn prediction challenge.

Our model based on survival ensembles outputs accurate predictions of {\em when} players churn, and provides information about the risk factors that affect the exit of players as well.
Additionally, the approach suggested in this paper not only gives us a list of possible churners, but also produces, for every player, a {\em survival} probability function that will let us know how the probability of churning 
is varying as a function of time.
This feature lets us distinguish various levels of loyalty profiles, upcoming, near-future and far-future churners, and the variables that influence this {\em survival} behavior (considering that a player is {\em alive} as long as she connects to the social game).

From this {\em survival} function, the median survival time is extracted and used as a life expectancy threshold. This feature lets us label players as being at risk of churning, take action beforehand to retain valuable players, and ultimately improve game development to enhance player satisfaction. 

To the best of our knowledge, we are the first to thoroughly model the prediction of churn by using a survival ensemble approach in the social games sector. 
Our model improves the accuracy, robustness and flexibility of traditional survival methods, like Cox regression, and has been developed with the goal of being usable in an operational business environment.

%-------------------------------------------------------------------------------
%
%-------------------------------------------------------------------------------
\section{Survival Ensemble Models}
\label{exp_desp}

\subsection{Survival Analysis}
Survival analysis consists of a set of statistical techniques traditionally used to predict lifetime expectancy of individuals in medical and biological research \cite{li2013survival, hougaard1999fundamentals,fleming2000survival}. 
This group of methods have also been applied in several industries to predict customer attrition, mainly in telecommunication \cite{lu2002predicting}, banking \cite{StepanovaT02} and insurance \cite{fuestimating}.

Survival analysis focuses on studying the {\em time} until an {\em event} of interest happens and its relationship with different factors. 
Originally in medical research, an event is the failure or death of an individual, however in our case it is the moment when a player leaves the game.
The {\em time-to-event} outcome is also known as {\em survival time}.

A fundamental characteristic of survival analysis is that data are {\em censored}. Censoring indicates that observations do not include complete information 
about the occurrence of the event of interest. It means that for a certain number of players, we do not know the time of event experience (because they did not experience it yet), i.e. 
measurements only contain information if the event occurs or not before a given time $t$.

The survival function $S(t)$, which is simply the likelihood that a player will survive at a certain time $t$, can be estimated through the non-parametric {\em Kaplan-Meier} 
estimator \cite{kaplan58}, where the churn probability can be computed directly from recorded censored survival times. 

If $k$ players churn during the period of time $T$ of study at different instants $t_1 < t_2 < t_3 \dots < t_k$
and, as churn occurrences are supposed to be independent of each other \cite{clark}, 
the probabilities of surviving in the game from one time to the next can be multiplied 
to obtain the {\em cumulative survival probability}:
\begin{equation}
S(t_j) = S(t_{j-1})\bigg(1 - \frac{d_j}{n_j} \bigg)
\label{km_est}
\end{equation}
where $S(0)=1$, with $n_j$ being the number of players alive before $t_j$, and $d_j$ being the number of events at $t_j$.
We will get as a result a step function that changes its value at the time of each churn.

Further analysis on this topic includes the presence of {\em competing risks} \cite{Prentice}. They belong to a special class of time-to-event models where there is more than one possible failure event. These alternative events can prevent the observation of the main event of interest. In this study, we focus on the loss of interest in a game, which is the main cause of churn. However, it can happen that a player stops playing the game because she loses her phone, or dies, which are considered as competing risks events.

Additional semi-parametric survival techniques, like the renowned regression method for censored observations, the {\em Cox proportional-hazards model} \cite{Cox1972, cox1984analysis, david1972regression}, or parametric methods (e.g. accelerated failure time models \cite{MarubiniValsecchi04}), are
valuable tools to investigate the impact of multiple covariates. The covariates or predictors are expected to be correlated with the player's reason for quitting the game. 

Following Cox proportional-hazards model, the estimated hazard for $k$ individual players and $p$ covariate vectors $x_k$ takes the form
\begin{equation}
h_k(t)=h_0(t) \exp{\big(\beta_1 x_{k,1} +  \dots + \beta_p x_{k,p}\big)},
\label{cox}
\end{equation}
where the hazard function $h_k(t)$ is dependent on the baseline hazard $h_0(t)$ and the features $\beta_p x_{k,p}$.
The Cox regression is not assumed to follow a particular statistical distribution. It is fitted based on the data and it solves 
the censoring problem by maximizing the partial likelihood. 

The Cox model and its extensions \cite{survival-book} allow regressions to work with censored data, and they permit an intuitive interpretation of the impact of the features. However, these 
techniques assume a fixed link between 
the output and the variables (assuming them additive and constant over time). This requires an explicit specification of the relationship by the researcher, and involves important efforts in terms of model selection and evaluation. In spite of their semi-parametric nature, these models present difficulties to scale with big data problems, and alternative {\em regularized} versions of Cox regression \cite{mittal2013high} have been proposed to amend this. Nevertheless, they are still based on restrictive assumptions that are not easy to fulfill.

In the parametric approaches, like the accelerated failure time models \cite{MarubiniValsecchi04}, the type of the distribution is previously determined (e.g. Weibull, lognormal, exponential). Though, these methods are suboptimal because it is uncommon that the data follow these specific distribution shapes.  

%Nevertheless, currently more flexible alternatives are available that solve these drawbacks. 

In the present paper, we address the drawbacks mentioned above by applying machine learning algorithms to censored data problems.

\subsection{Survival Trees and Ensembles}
\label{theory}
\subsubsection{Decision Trees}
Originally presented in \cite{morgan1963problems}, decision trees became popular in the 1980s, when the most 
relevant algorithms for {\em Classification and Regression Trees} (CART) 
were introduced by \cite{breiman1984classification,quinlan1986induction,salzberg1994c4}.\\
Classification and regression trees are non-parametric techniques where the basic idea is to split the feature space recursively, to group subjects with homogeneous characteristics 
and to separate those with bigger differences based on the outcome of concern. In order to perform the nodes classification and maximize homogeneity within the nodes, 
a measure called {\em impurity} must be minimized. Common examples of impurity measure are cross-entropy or sum of squared errors.
For example, considering a binary split and given a continuous variable $X$, the split can be performed if $X \leq d$ is fulfilled, with $d$ being a constant.

\subsubsection{Survival trees}
Survival trees are constructed as a set of binary trees that grow by recursive partitioning of the sample space $\chi$, where the $q_i$ tree nodes are 
subspaces of $\chi$. The tree splitting starts in the root node, which concentrates 
all the data. Based on a survival statistical criterion, such as the cumulative hazard function or Kaplan-Meier estimates, the root node is then divided into two daughter nodes. The principle for partitioning these 
two branches is to maximize the survival difference between two groups of individuals, which are compressed in the two daughter nodes, maximizing the homogeneity among nodes, based on survival experience.

The first idea of using tree-based methods for censored data was initially introduced in \cite{ciampi1981approach} and \cite{marubini1983prognostic}. The first {\em survival tree} 
as we know was presented in \cite{gordon1985tree}, where a Kaplan-Meier estimator survival function was computed at every node as a discrepancy measure using Wasserstein metrics. For a comprehensive review about different types of survival trees, check \cite{bou2011review}.

The best split is achieved by exploring all combinations, considering all the $x_i$ predictor variables and all the possible splits, in order to maximize the survival difference.
This way, subjects with similar survival characteristics are grouped together. As long as the tree grows, the difference between branches increases, and individuals are gathered in nodes with more homogeneous groups in terms of survival behavior.

Despite being a powerful classification tool which is able to model censored data, employing a single tree can produce instability in its predictions. It means that if small changes in data arise, the prediction 
can differ among computations (the divergences are mainly related with the prediction of risk factors) \cite{Kretowska2014}. This drawback will be fixed if we execute an ensemble of them, instead of using one single tree.

\subsubsection{Survival ensembles}
\label{survEn}
Using an ensemble of models, instead of a single one, is an accurate prediction tool firstly suggested by \cite{breiman1996bagging, breiman2001random} with the well-known {\em random forest}. 
Ensembles of tree-based models achieve outstanding predictions in real-world applications \cite{zhang2012ensemble}. 

Survival forests are ensemble-based learning methods where the underlying algorithm is a kind of survival tree. 
A survival ensemble lies in growing a set of survival trees, instead of a single one.
The two main survival ensemble techniques are
{\em random survival forest}, presented in \cite{ishwaran2008random}, and
{\em conditional inference survival ensembles}, developed by \cite{Hothorn06unbiasedrecursive}, based on their previous work introduced in \cite{hothorn2004bagging,hothorn2006survival}.
%like in the work proposed by \cite{hothorn2006survival} where Kaplan-Meier survival functions are aggregated for prediction.

The conditional inference survival ensembles is the method chosen for the predictions shown in Section \ref{experiments}.
The conditional inference survival ensemble technique uses a weighted Kaplan-Meier function based on the measurements used for the training. The ensemble survival function \cite{mogensen2012evaluating} can be summarized by
\begin{equation}
S^{conditional}(t|x_i) = \prod \bigg(1 - \frac{\sum_{n=1}^{N} T_n(dt,x_i)}{\sum_{n=1}^{N} Q_n(t,x_i)} \bigg)
\label{condEnsem}
\end{equation}
where $n$ indicates the number of trees within the ensembles, with $n = 1, \cdots, N$, and $x_i$ being the covariates.
Therefore, in the node where $x_i$ is located, $T_n$ accounts for the 
uncensored events until time $t$, and $Q_n$ counts 
the number of individuals at risk at time $t$.
%In \cite{ishwaran2008random}, another survival ensemble approach, called {\em random survival forest}, is proposed. Models built using random survival forests are based on Nelson-Aalen estimates instead of using Kaplan-Meier estimates, which is the main difference with the proposed approach.
Moreover, conditional inference survival ensembles introduces additional weight to the nodes where there are more subjects at risk. It uses linear rank statistics as splitting criterion to grow the trees. 

In contrast, random survival forests \cite{ishwaran2008random} are based on Nelson-Aalen estimates (instead of using Kaplan-Meier estimates). The maximum of the log-rank statistical test is used in every node as split criterion, which leads to biased results in favor of covariates with many splits. 
%which is the main difference with the proposed approach.

Conditional inference survival ensembles is a promising approach to deal with the censoring nature of churn prediction. It is a flexible method compared to the traditional statistical Cox regression model and it solves the instability that is present in survival trees. In the selected method for the churn study, overfit is not present in its estimates and provides robust information about the variable importance. This fixes the 
 random survival forest problem \cite{wright2016random} of being biased towards predictors with many splits or missing data.

%-------------------------------------------------------------------------------
%
%-------------------------------------------------------------------------------
  
\section{Dataset}
\label{experiments}
%The major goal of survival models is to use information about the lifetime history of players and their behavior in the game, 
%i.e. predictors, covariates or features, to predict future time-to-event outcomes. 
%In this study, we use the survival forest algorithm, developed by \cite{Hothorn06unbiasedrecursive}, based on the conditional inference framework. The 
%event of interest of our model is whether the players are going to leave the game, and if so, in how many days.

% 
% **Define churn: open problem. Include a plot of time series session and purchase.
% \begin{figure}[ht!]
%   \centering
% %  (a)\includegraphics[height=6cm]{soil_no}
% %  (b)\includegraphics[height=6cm]{soil_yes}
%   \includegraphics[height=2cm]{seriesTime.png}\\
%   \includegraphics[height=2.5cm]{seriesPurchase.png}
% %
%   %  (b)\includegraphics[height=4.3cm]{images/O_channel2_4h.png}
% \caption{Time Series of time spent playing (upper panel) and time series in-app purchases (below) by an specific HVP player of {\em Age of Ishtaria}.}
% \label{KM}
% \end{figure}

%The data collected covers 493 days from October 2014 to February 2016.
%Several churn predictors or risk factors were investigated.
We collected data from a major mobile social game between October 2014 and February 2016. Several churn predictors or risk factors were investigated.

%We aim at building a game-independent churn prediction model so that it can be applied to other games. Therefore 
We investigated mainly game-independent features, i.e. features that are not related to the game mechanics and can be measured in any game. This allows us to build a game-independent churn prediction model that can be applied to other games.

Additionally, we want to implement our model in a data science product running in an operational business environment. Thus, the feature selection takes into account limitations in terms of memory and processing capabilities that might not be considered in a pure research environment.

\begin{itemize}
 \item Player attention: the time component of the player accessing the game.
 \begin{itemize}
  \item Time spent per day in the game, including averages over the first weeks and moving average over the last weeks.
  \item Lifetime: number of days since registration until churn, in case the player churns.
  \end{itemize}

 \item Player loyalty: the frequency of the player access to the game. %%
  \begin{itemize}
 \item Number of days with at least one playing session.
 \item Loyalty index: ratio of number of days played, divided by lifetime.
 \item Days from registration to first purchase.
 \item Days since last purchase.
 \end{itemize}

\item Playing intensity: the quality of the playing sessions, i.e. how a player interacts with the game.
 \begin{itemize}
 \item  Number of actions.
  \item Number of sessions.
  \item Number and amount of in-app purchases.
\item Action activity distance: Euclidean distance between the average number of actions over the lifetime and the average number of actions over the last days.
  \end{itemize}
\item Player level: the value of this variable and its evolution depends on the game. However, the concept of level is present and measurable in the majority of games, and can be then considered as a game-independent predictor that can be used in our model and applied to most other mobile social games.
 \end{itemize}

We investigated some game-dependent features, that we ultimately did not keep in our model, such as:
\begin{itemize}
  \item Participation in a {\em guild}. Guild is a social feature, sometimes called {\em union} or {\em clan}, specific to some social games, allowing to play in collaboration with other players.
  This predictor turned to be inapplicable to our problem as the whales, who are the focus of this study, have an homogeneous behavior in terms of participation to the social features of the game.
\item Measure of number of actions by category (shop, battle, mission, ...). This is 
specific to the game studied. Though, it does not bring more relevant information than the 
higher-level and game-independent measure of the total number of actions. 
  \end{itemize}
 
%-------------------------------------------------------------------------------
%
%-------------------------------------------------------------------------------
\section{Modeling}
\label{sec3}
\subsection{Churn definition}
As it was explained in Section \ref{intro}, the definition of churn in F2P games is not straightforward.
In this study, we consider that a player has churned if she does not connect to the game for 10 consecutive days.
Our measurements confirm that the whales who went through a period of 10 days of inactivity become mostly inactive: they either permanently exit the game or their activity becomes neglectable.
Indeed the purchase activity of whales after coming back to the game following a period of 10 days of inactivity represents only 1.4$\%$ of the revenues generated by this category of players.

Traditional churn analysis focuses on predicting whether or not a user is going to exit the game, i.e. the response is a binary variable: yes or no. However, with this approach, we do not know when a player is going to stop connecting to the game. Conventional churn prediction is solved from a static point of view, a binary classification problem.

Our work focuses on {\em when} churn will happen. We model the churn behavior from the perspective of survival analysis, and we treat the prediction of churners as a censored data problem where the outcome of our model is the continuous time - the {\em time-to-exit} the game. We have used the 
algorithm of survival ensembles within the conditional inference framework from \cite{Hothorn06unbiasedrecursive}, presented in Section \ref{survEn}.
This study uses a learning sample of $n=2500$ whales. 

\begin{figure}[t]
  \centering
%  (a)\includegraphics[height=6cm]{soil_no}
%  (b)\includegraphics[height=6cm]{soil_yes}
  \includegraphics[width=\columnwidth]{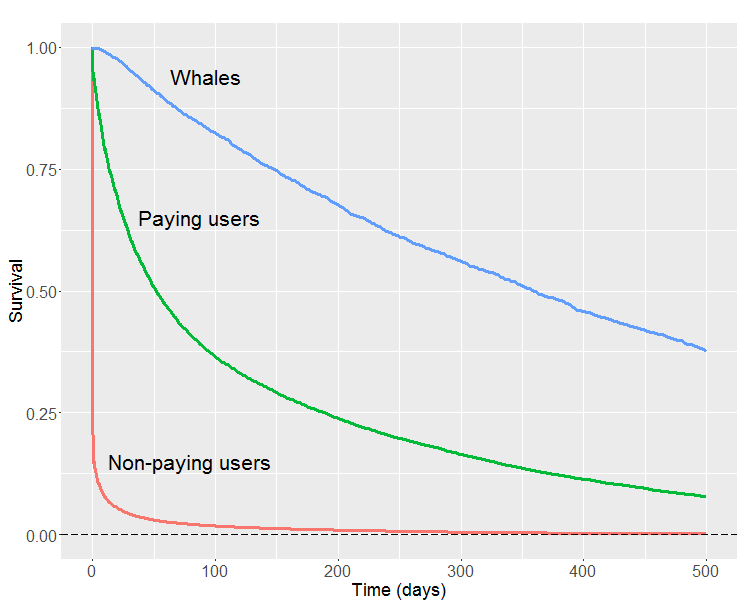}\\
%  (b)\includegrawidth=\textphics[height=4.3cm]{images/O_channel2_4h.png}
\caption{Cumulative survival probability (Kaplan-Meier estimates) as a function of time, in days. The whole dataset is considered, and the results are stratified by whales (high-value players), paying users, and non-paying users.}
\label{KM}
\end{figure}

\subsection{Kaplan-Meier estimates}
We visualize the churn problem by plotting Kaplan-Meier (K-M) survival curves stratified by whales, 
normal paying users, and non-paying users. In order to perform the K-M survival analysis, we take a sample of 1.500.000 players.

\begin{figure*}[ht!]
  \centering
%  (a)\includegraphics[height=6cm]{soil_no}
%  (b)\includegraphics[height=6cm]{soil_yes}
  \includegraphics[width=\textwidth]{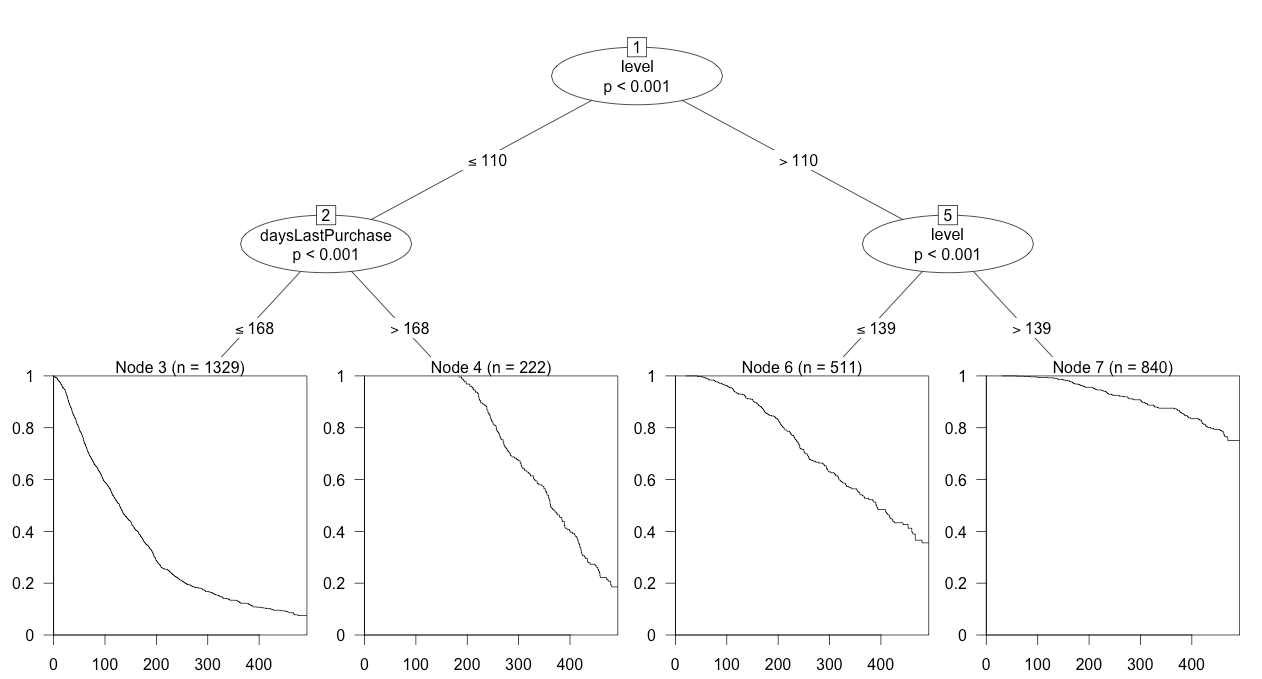}
%  (b)\includegraphics[height=4.3cm]{images/O_channel2_4h.png}
\caption{Conditional inference survival tree partition with the estimated Kaplan-Meier survival curve. The K-M estimates represent the survival probability (in days), which characterizes the players placed in every terminal node.}
\label{survTree}
\end{figure*}

Fig. \ref{KM} provides a graphical representation of the K-M survival curves for different kind of players based on their paying behavior, distinguishing among whales (high-value players), paying users, and non-paying users. Fig. \ref{KM} shows different survival patterns for each group. The estimated survival for non-paying users is much lower than the one of paying users (including both whales and non-whales paying users). Approximately 80$\%$ of the non-playing players have churned the first day they connected to the game. This contrasts with the 20$\%$ churn rate of the whales after 100 days.

% \begin{table}
% \centering
% \caption{Features included in the two churn models considered.}
% \begin{tabular}{|c|c|} \hline
% Model& Features\\ \hline
% Binary problem & loyalty index\\ %\hline
% &lifetime \\ 
% &level \\ 
% &average playing time first 15 weeks \\ 
% &days since last purchase\\ 
% &days to first purchase\\ 
% & action activity distance\\ \hline
% Censored data& loyalty index \\ %\hline
% &level \\ 
% &last purchase amount \\ 
% &days since last purchase,\\ 
% &days to first purchase\\ 
% &amount first purchase\\ 
% &average playing time first 15 weeks\\ 
% & action activity distance\\ \hline
% \end{tabular}
% \label{tabla1}
% \end{table}

\subsection{Churn model as a censored data problem}

In the present work, the authors propose conditional inference survival ensembles \cite{Hothorn06unbiasedrecursive} to model game churn.

Survival ensembles with 1000 conditional inference trees are used as a base learner to predict the exit time of whales from the game.
Fig. \ref{survTree} shows how conditional inference trees work. It illustrates a simple partition with two terminal nodes. In each terminal node,
a Kaplan-Meier survival curve represents the group of players included in the node classification. In this example we can observe the differences between
the survival profiles that characterize every node. In Fig. 
\ref{survTree}, the root node variable is the last level 
the player reached in the game. Two daughter nodes partitions grow from it: one also based on the level and another based on the number of days since 
the player did the last in-app purchase {\em daysLastPurchase}.

\begin{figure}[ht!]
  \centering
%  (a)\includegraphics[height=6cm]{soil_no}
%  (b)\includegraphics[height=6cm]{soil_yes}
  \includegraphics[width=\columnwidth]{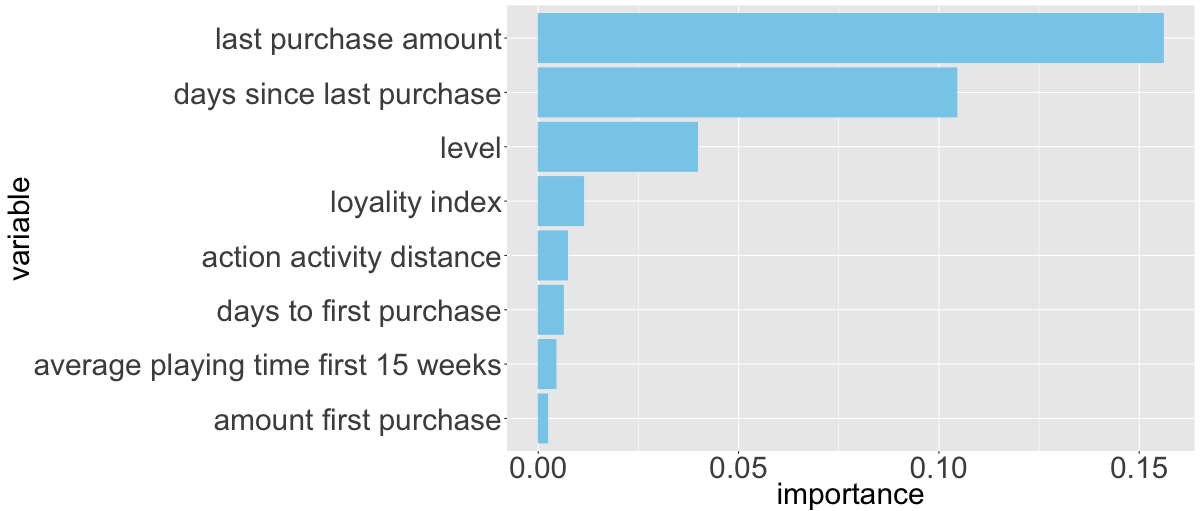} \\
  \includegraphics[width=\columnwidth]{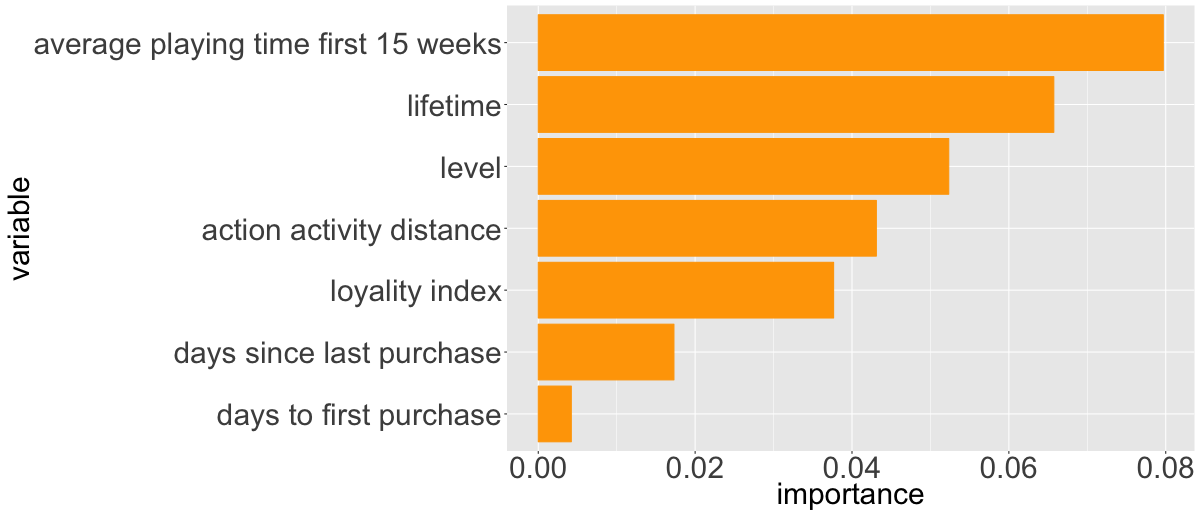} 
%  (b)\includegrawidth=\textphics[height=4.3cm]{images/O_channel2_4h.png}
\caption{Variable importance (relative ranking of significant factors) to predict time-to-churn with conditional inference survival ensemble model in the upper panel, and for 
the binary output in the lower panel.}
\label{importance}
\end{figure}

%The overall survival time, together with the churn censoring indicator, is the outcome of this model. Different kind of features, both game independent and game dependent, 
The overall survival time is the outcome of this model. Fig. \ref{importance} summarizes the most significant 
predictors included in the survival ensemble model for right-censored observations. The variable importance is computed 
using the integrated Brier score (IBS) \cite{graf1999assessment}, and the feature selection is performed based on it. Other survival ensemble methods, like \cite{ishwaran2008random}, are not as robust as the technique employed in this work \cite{Hothorn06unbiasedrecursive} in terms of variable selection and therefore in terms of computation of variable importance. The variable importance is 
normally biased in favor of the predictors with many splits. Conditional inference survival ensembles are constructed based on unbiased trees, avoiding 
this problem \cite{hothorn2015package}.

%The results reported here are based on... 
\begin{figure*}[ht!]
  \centering
%  (a)\includegraphics[height=6cm]{soil_no}
%  (b)\includegraphics[height=6cm]{soil_yes}
  \includegraphics[width=0.49\columnwidth]{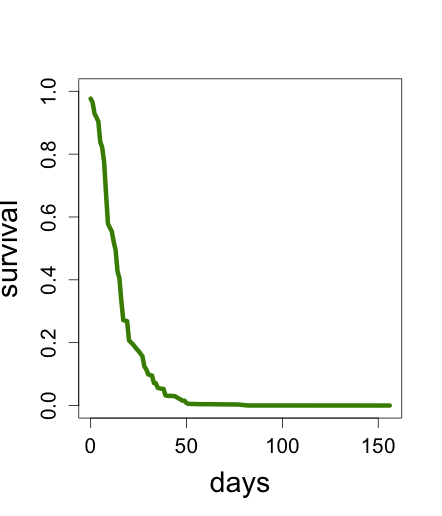}
  \includegraphics[width=0.49\columnwidth]{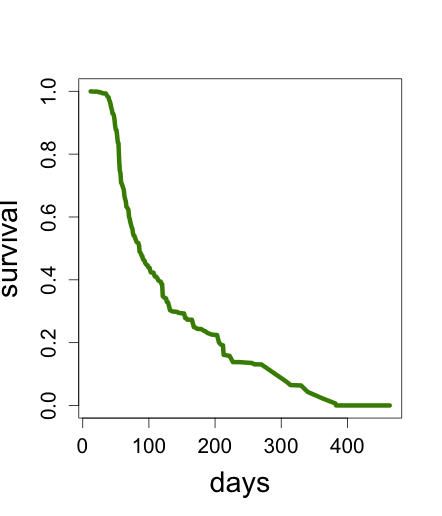}
  \includegraphics[width=0.49\columnwidth]{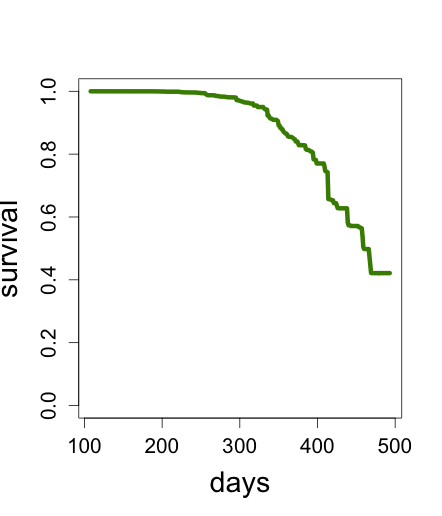}
  \includegraphics[width=0.49\columnwidth]{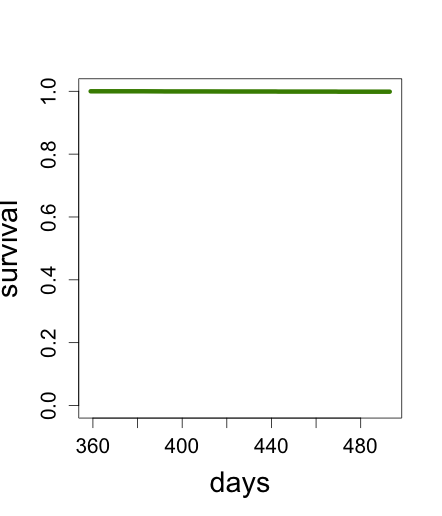}

  %  (b)\includegraphics[height=4.3cm]{images/O_channel2_4h.png}
\caption{Predicted Kaplan-Meier survival curves as a function of time (days) for new players. Predictions performed with conditional 
inference models for survival ensembles.}
\label{player_profile}
\end{figure*}
The resulting prediction of this model contains, for each player, a survival function indicating the probability of churn as a function of time since the registration in the game.
Fig. \ref{player_profile} illustrates a sample of four Kaplan-Meier survival functions for four new players. In Fig. \ref{player_profile}, we can observe 
the probability of churn for every single player (y-axis) as a function of time in days (x-axis). In this example, we distinguish different player profiles and survival behaviors:  
\begin{itemize}
 \item The two first plots starting from the left show the survival probability curves of two players who are going to churn soon.
\end{itemize}
%  \begin{enumerate}[label=(a)]
%   \item short life expectancy probability
%   \item medium life expectancy probability
%   \item long life expectancy probability
%  \end{enumerate}
\begin{itemize}
 \item The third plot starting from the left shows the survival probability curve of a player who is expected to churn but not in the near future.
 \item The last plot starting from the left shows the survival probability curve of a very loyal player.
 \end{itemize}
For every player, a different survival function will be computed as a result of our model.
 
Fig. \ref{player_profile} 
highlights the capability of our model to classify and predict loyalty for every player, taking into account the temporal dimension. Additionally, the median survival time, which is the time when the percentage of surviving in the game is 50$\%$, is used as a time threshold to categorize a player as being at risk of churning.

\subsection{Model validation}
\begin{figure*}[ht!]
  \centering
%  (a)\includegraphics[height=6cm]{soil_no}
%  (b)\includegraphics[height=6cm]{soil_yes}
  \includegraphics[width=0.95\columnwidth]{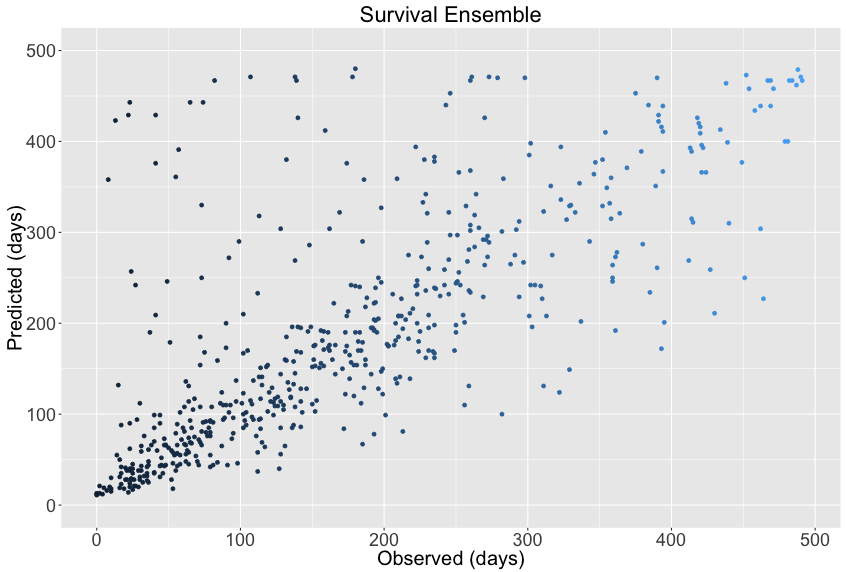}
  \includegraphics[width=0.95\columnwidth]{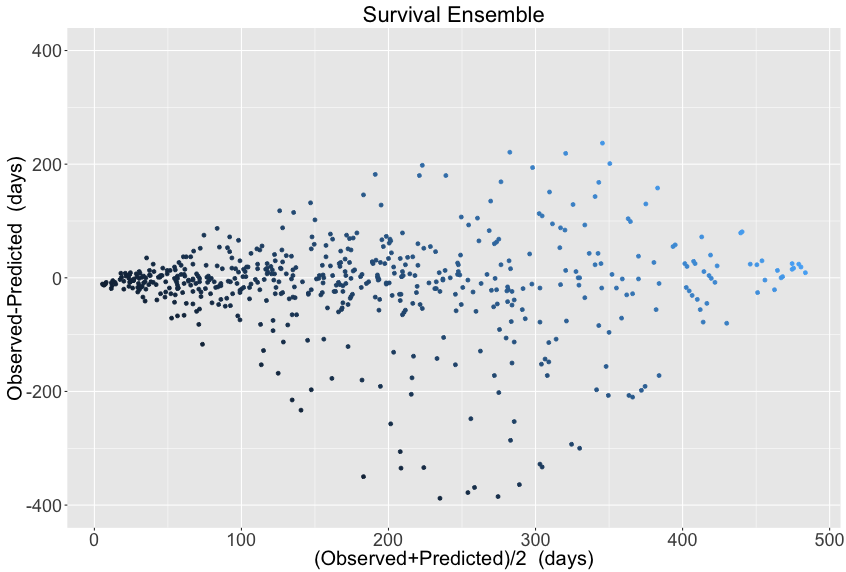}\\
  \includegraphics[width=0.95\columnwidth]{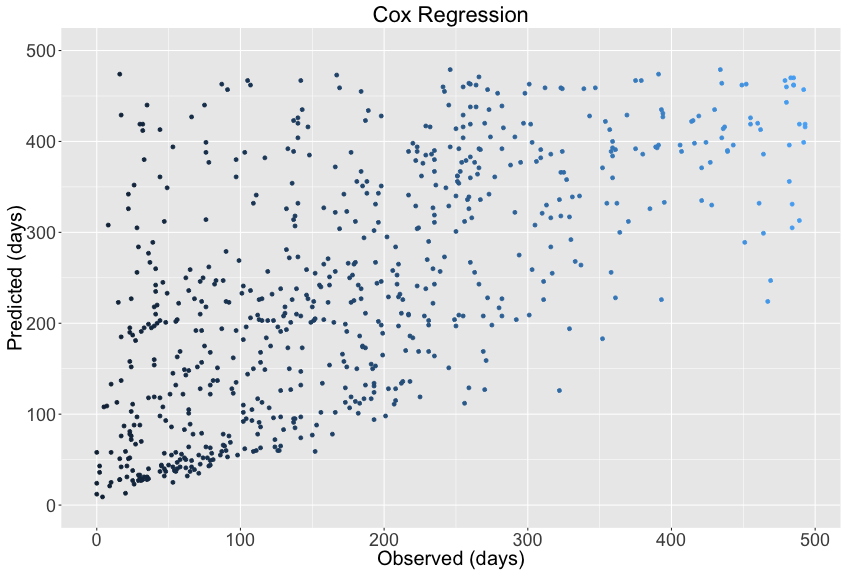}
  \includegraphics[width=0.95\columnwidth]{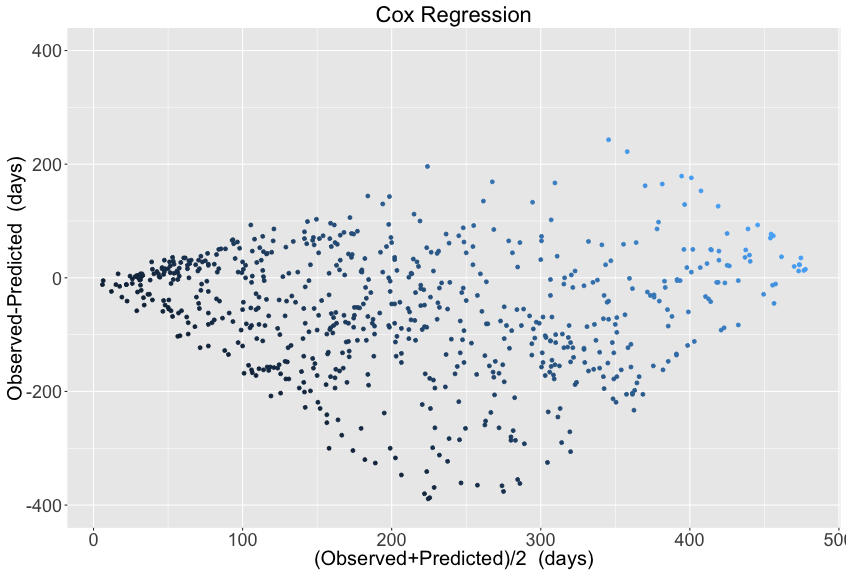}
  %  (b)\includegraphics[height=4.3cm]{images/O_channel2_4h.png}
\caption{Scatterplot (left) and  mean-difference plot (right) of observations and predictions of median survival times. The dark blue 
dots correspond to shorter lifetimes (in days) of players, soft blue dots reflect players with longer lifetimes. Upper panel evaluates the 
survival ensemble results, and lower panel compares the Cox regression analysis.}
\label{scatterplot}
\end{figure*}

Because of the nature of censoring, the standard methods of visualizing and evaluating prediction performances are not suitable \cite{mogensen2012evaluating}. Fig. \ref{scatterplot} shows 
the fit of the proposed conditional inference survival ensemble method and the selected Cox regression (using the same predictors). The conditional inference survival ensemble model 
exhibits a reasonable agreement between measured and predicted survival times, both in the scatterplot and the mean-difference plot evaluation. We can observe in the lower 
plots that Cox regression performs worse than the ensemble model in terms of predictive ability. As it can be observed in Fig. \ref{scatterplot}, there is a higher 
concentration of data at the beginning of the study. This is due to the fact that we work with 
censored data and do not follow a normal distribution. Hence, the longer the time of study grows, the less information we have, as there are many whales who have not experienced the {\em event} 
yet because they are still connecting to the game. This evidence is reflected in the cumulative survival distribution for whales shown in Fig. \ref{KM}. 
Thus, as long as the censoring rate grows, the prediction capability diminishes.
%-------------------------------------------------------------------------------
%
%-------------------------------------------------------------------------------
\begin{figure}[ht!]
  \centering
%  (a)\includegraphics[height=6cm]{soil_no}
%  (b)\includegraphics[height=6cm]{soil_yes}
  \includegraphics[width=\columnwidth]{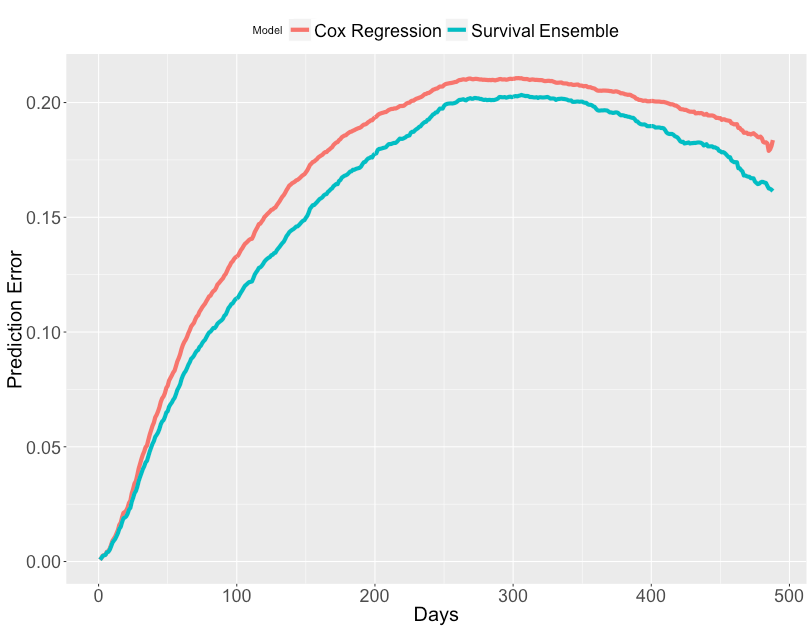}\\
  %  (b)\includegraphics[height=4.3cm]{images/O_channel2_4h.png}
\caption{1000 bootstrap cross-validation error curves for the survival ensemble model and Cox regression.}
\label{predErrors}
\end{figure}

%-------------------------------------------------------------------------------
%
%-------------------------------------------------------------------------------

%-------------------------------------------------------------------------------
%
%-------------------------------------------------------------------------------

Fig. \ref{predErrors} depicts the cumulative prediction error curve for the survival ensemble and the Cox regression model. The integrated Brier score is an evaluation measure developed for 
survival analysis \cite{mogensen2012evaluating, graf1999assessment}. We use it to establish the summary of the error estimation for the two survival-time analysis outputs. The error evaluation has been performed based on 
bootstrap cross-validation with replacement. This technique estimates the prediction error splitting the measurements in many bootstrap training and test samples. Then, the models are trained and tested with multiple sets of bootstrap samples. Fig. \ref{predErrors} exhibits the bootstrap cross-validated prediction error curves for 1000 samples. 
\begin{table}
	\centering
	\caption{Prediction error score for the survival models with Integrated Brier score (IBS)}
	\begin{tabular}{|c|c|} \hline
		Model                               & IBS\\ \hline
		Survival Ensemble            &  0.158\\ 
		Cox regression &  0.169\\ 
		Kaplan-Meier                 & 0.199\\ \hline
		 
	\end{tabular}
	\label{BrierTable}
\end{table}

Fig. \ref{predErrors} basically supports what Fig. \ref{scatterplot} shows, as the ensemble-based approach improves accuracy over the Cox model, cf. Table \ref{BrierTable}. The prediction error function reaches the maximum at the median survival time of 304 days and 306 days, for the Cox regression (error value of 0.21) and ensemble model (error value of  0.20), respectively.

Additional validation tests have been performed to compare the accuracy of the two models. A paired t-test (Welch Two Sample t-test) \cite{national2001engineering} is used to estimate whether the prediction ability of a model is statistically significant from another. The t-test has been performed using a confidence interval of $95\%$. According to the t-test, survival ensemble model is statistically significant, i.e. $p$-value $\leq 0.05$ . We obtain the following values: $t =  3.56$ and $p= 0.00039$.

\section{Comparison with other model approaches}
\label{compa}

We include in this section a binary classification model of churners. 
Although we think that modeling churn as a censoring problem is the adequate approach, the binary prediction perspective also brings us interesting information. The binary response model provides useful insight for a very short-term prediction. It is easy to interpret and to implement.

Although we use the same algorithm of conditional inference ensembles, the outcome differs. A binary variable denoting if a player churns or not is the response of the classification model, i.e. yes or no.
We trained the binary model with several sets of features to obtain the final list of attributes shown in Fig. \ref{importance}. We highlight the 
contrasting results obtained during the evaluation of the variable impact between the survival model and the binary classification. It reflects the nature of different ways of modeling and therefore of the prediction results.

A comparison study with other binary classification methods was performed in order to support 
the results obtained with the binary approach of the churn analysis. 
For this study, we select several algorithms as binary classificators: SVM, naive bayesian classifier
and a decision tree. A detailed and complete explanation about the techniques used here can be found in \cite{hastie_09_elements-of.statistical-learning}.

The fit of the ensemble is summarized in Table \ref{tablaNOMBRE}, where we compare our results with other classification methods.
It indicates a good agreement between observed and predicted churners with an AUC (area under ROC curve) of 0.96. Although the other techniques 
also perform very accurately, they possess some drawbacks.
SVM also have a high score in terms of AUC, but they are considered as {\em black boxes} because it requires significant effort to extract the relationship between the input variables and the output \cite{Kretowska2010}.

\begin{table}
	\centering
	\caption{AUC (area under ROC curve) performance of different binary classification algorithms}
	\begin{tabular}{|c|c|} \hline
		Model                               & AUC\\ \hline
		Survival Ensemble            & 0.960\\ 
		Support Vector Machines & 0.940\\ 
		Naive Bayesian                 & 0.900\\ 
		Decision Tree                    & 0.934\\ \hline
	\end{tabular}
	\label{tablaNOMBRE}
\end{table}

The techniques applied above are powerful tools to solve regression or classification problems. However, in their 
original, form they cannot handle the assimilation of information from censored data. Hence, in order to apply these methods for 
survival analysis responses, an adequate modification of the algorithm or a proper transformation of the data must be performed beforehand.

% \begin{figure}[ht!]
%   \centering
% %  (a)\includegraphics[height=6cm]{soil_no}
% %  (b)\includegraphics[height=6cm]{soil_yes}
%   \includegraphics[width=\columnwidth]{comprDef.png}
% %  (b)\includegraphics[height=4.3cm]{images/O_channel2_4h.png}
% \caption{ROC performance comparison between several models including SVM, naive bayesian classifier, decision tree and conditional survival ensemble. All 
% the models were applied with the same sample of data from {\em Age of Ishtaria}.}
% \label{comp}
% \end{figure}

%\vspace{0.2cm}

%%TODO
% % ADDED TABLE %%%%%%%%%%%%%%%%%%%%%%%%%%%%%%%%%%%%%%%%%
% \begin{table}
% 	\centering
% 	\caption{Features included in the two churn models considered.}
% 	\begin{tabular}{|c|ccc|ccc|} \hline
% 		    Thr.    & &Train& & &Test& \\ \hline
%             days      & acc. & rmse & mape & acc. & RMSE & MAPE \\ \hline
%               5         & 25.54      &  &  & 17.44 &  &  \\ 
%               10       & & & & & &  \\ 
%               30       & & & & & &  \\ 
%               50       & & & & & &  \\ 
%               75       & & & & & &  \\ 
%              100      & & & & & &  \\ 	\hline
% 	\end{tabular}
% 	\label{tabla1}
% \end{table}
% %%%%%%%%%%%%%%%%%%%%%%%%%%%%%%%%%%%%%%%%%%%%%%%%%%

\section{Summary and Conclusion}
\label{con_6}
% With a percentage of churn that ranges between 60$\%$ and
% 85$\%$ in the first day, retaining players is one of the most crucial challenges in mobile social games.
% Additionally, the cost of acquiring a new player can significantly exceed the cost of retaining an existing one.
% 
% TODO put again later
% 

The focus of this research is to find an appropriate technique to model player churn, which has been an open problem within the community. Furthermore, this work presents steps towards the challenging goal of understanding the most valuable players in social games.

The authors propose the application of a state-of-the-art algorithm: conditional inference survival ensembles \cite{Hothorn06unbiasedrecursive}, to predict the time-to-churn and the survival probability of players in social games in terms of game lifetime.

We look for a method that is able to make predictions in an operational business environment and 
that easily adapts to different kinds of games, players, and therefore distribution of the data. This is the main motivation: we need a flexible technique that does not 
require a previous manipulation of the data and that is able to deal efficiently with the temporal dimension of the churn prediction problem.
Conditional inference survival ensembles were evaluated to this purpose and compared with traditional survival methods like Cox regression.

Conditional inference survival ensembles provided more accurate and more stable prediction results than traditional approaches. The proposed method is unbiased, does not overfit \cite{Hothorn06unbiasedrecursive}, and provides us with robust information about the risk factors that influence players to abandon the game.

The predictions we have obtained provide the business users and game developers with useful and easy-to-interpret player information. The results 
directly impact the game business, improving the knowledge about whales behavior, discovering new playing patterns as a function of time, and classifying social gamers by risk factors of churn.

Further on-going work in this direction is the improvement of the accuracy in the prediction of the time-to-churn for players who stay longer in the game. To achieve this, 
we will continue researching significant features to discover new playing patterns. A promising direction would be to study predictors based on more complex measures of the social activity than the one used in this study.

\vspace{0.2cm}
\section{Software}
\label{con_7}
All the analysis were performed with R version 3.1.2 for Linux, using the following packages from CRAN: {\em party} 
version 1.0 \cite{hothorn2010party,hothorn2015package}, {\em survival} 
version 2.38 \cite{therneau2015package}, {\em ROCR} 1.0 \cite{sing2005rocr,sing2007package}, {\em Core Learn} version 
1.47 \cite{robnik2012corelearn,robnik2013package} and {\em cvTools} version 0.3.2 \cite{alfons2012cvtools}.

\section*{Acknowledgements}
\label{con_8}
We thank our colleague Sovannrith Lay for helping us to collect the data and his support during this study. We also thank Thanh Tra Phan for the careful review of the article.

\bibliographystyle{abbrv}
\bibliography{churnsig}

\end{document}